\documentclass{article}

\usepackage{arxiv}

\usepackage[utf8]{inputenc} 
\usepackage[T1]{fontenc}    
\usepackage{hyperref}       
\usepackage{url}            
\usepackage{booktabs}       
\usepackage{amsfonts}       
\usepackage{nicefrac}       
\usepackage{microtype}      
\usepackage{lipsum}		
\usepackage{graphicx}
\usepackage[numbers]{natbib}
\usepackage{doi}
\usepackage{caption}
\usepackage{threeparttable}
\usepackage{placeins}
\usepackage{float} 
\usepackage[affil-it]{authblk} 
\usepackage{hyperref} 

\title{Incorporating Anatomical Awareness for Enhanced Generalizability and Progression Prediction in Deep Learning-Based Radiographic Sacroiliitis Detection}

\author[1,2]{\textbf{Felix J. Dorfner} \textsuperscript{*}\textsuperscript{$\dagger$}}
\author[1]{Janis L. Vahldiek}
\author[1]{Leonhard Donle}
\author[3]{Andrei Zhukov}
\author[1]{Lina Xu}
\author[1]{Hartmut Häntze}
\author[4]{Marcus R. Makowski}
\author[5,6,7]{Hugo J.W.L. Aerts}
\author[3]{Fabian Proft}
\author[3]{Valeria Rios Rodriguez}
\author[3]{Judith Rademacher}
\author[3]{Mikhail Protopopov}
\author[3]{Hildrun Haibel}
\author[1]{Torsten Diekhoff}
\author[3]{Murat Torgutalp}
\author[4]{Lisa C. Adams}
\author[3,8]{\textbf{Denis Poddubnyy}\textsuperscript{$\ddagger$}}
\author[3,4,5,9]{\textbf{Keno K. Bressem}\textsuperscript{$\ddagger$}}

\affil[1]{Department of Radiology, Charité - Universitätsmedizin Berlin corporate member of Freie Universität Berlin and Humboldt Universität zu Berlin, Hindenburgdamm 30, 12203 Berlin, Germany}
\affil[2]{Athinoula A. Martinos Center for Biomedical Imaging, Massachusetts General Hospital and Harvard Medical School, 149 Thirteenth St, Charlestown, MA 02129, USA}
\affil[3]{Department of Gastroenterology, Infectious Diseases and Rheumatology (incl. nutrition medicine), Charité - Universitätsmedizin Berlin corporate member of Freie Universität Berlin and Humboldt Universität zu Berlin, Hindenburgdamm 30, 12203 Berlin, Germany}
\affil[4]{Department of Diagnostic and Interventional Radiology, Faculty of Medicine, Technical University of Munich, Ismaninger Strasse 22, 81675, Munich, Germany}
\affil[5]{Artificial Intelligence in Medicine (AIM) Program, Mass General Brigham, Harvard Medical School, 221 Longwood Avenue, Boston, MA 02115, USA}
\affil[6]{Departments of Radiation Oncology and Radiology, Dana-Farber Cancer Institute and Brigham and Women’s Hospital, 221 Longwood Avenue, Boston, MA 02115, USA}
\affil[7]{Radiology and Nuclear Medicine, CARIM \& GROW, Maastricht University, Minderbroedersberg 4-6, 6211 LK Maastricht, Netherlands}
\affil[8]{Department of Epidemiology, German Rheumatism Research Center, Charitéplatz 1, 10117 Berlin, Germany}
\affil[9]{Department of Radiology and Nuclear Medicine, German Heart Center Munich, Munich, Germany}

\date{}

\renewcommand{\headeright}{}
\renewcommand{\shorttitle}{Anatomical Awareness in Sacroiliitis Detection}

\hypersetup{
pdftitle={Incorporating Anatomical Awareness for Enhanced Generalizability and Progression Prediction in Deep Learning-Based Radiographic Sacroiliitis Detection},
pdfsubject={Pelvic X-Ray, Artificial Intelligence, Deep Learning, Radiology, Rheumatology, axial Spondyloarthritis},
pdfauthor={Felix J. Dorfner, Janis L. Vahldiek, Leonhard Donle, Andrei Zhukov, Lina Xu, Hartmut Häntze, Marcus R. Makows, Hugo J.W.L. Aerts, Fabian Proft, Valeria Rios Rodriguez, Judith Rademacher, Mikhail Protopopov, Hildrun Haibel, Torsten Diekhoff, Murat Torgutalp, Lisa C. Adams, Denis Poddubnyy, Keno K. Bressem},
pdfkeywords={Pelvic X-Ray, Artificial Intelligence, axial spondyloarthritis, Rheumatology, Radiology},
}

\begin{document}
\maketitle
\textsuperscript{*}First author.\\
\textsuperscript{$\dagger$}Corresponding author: felix.dorfner@charite.de \\
\textsuperscript{$\ddagger$}These authors contributed equally as last authors.

\abstract{\textbf{Purpose:} To examine whether incorporating anatomical awareness into a deep learning model can improve generalizability and enable prediction of disease progression. 
 
\textbf{Methods:} This retrospective multicenter study included conventional pelvic radiographs of 4 different patient cohorts focusing on axial spondyloarthritis (axSpA) collected at university and community hospitals. The first cohort, which consisted of 1483 radiographs, was split into training (n=1261) and validation (n=222) sets. The other cohorts comprising 436, 340, and 163 patients, respectively, were used as independent test datasets. For the second cohort, follow-up data of 311 patients was used to examine progression prediction capabilities. Two neural networks were trained, one on images cropped to the bounding box of the sacroiliac joints (anatomy-aware) and the other one on full radiographs. The performance of the models was compared using the area under the receiver operating characteristic curve (AUC), accuracy, sensitivity, and specificity. 

\textbf{Results:} On the three test datasets, the standard model achieved AUC scores of 0.853, 0.817, 0.947, with an accuracy of 0.770, 0.724, 0.850. Whereas the anatomy-aware model achieved AUC scores of 0.899, 0.846, 0.957, with an accuracy of 0.821, 0.744, 0.906, respectively. The patients who were identified as high risk by the anatomy aware model had an odds ratio of 2.16 (95\% CI: 1.19, 3.86) for having progression of radiographic sacroiliitis within 2 years. 

\textbf{Conclusion:} Anatomical awareness can improve the generalizability of a deep learning model in detecting radiographic sacroiliitis. The model is published as fully open source alongside this study.}

\keywords{Pelvic X-Ray, Artificial Intelligence, Axial Spondyloarthritis, Rheumatology, Radiology}
\endabstract

\section{Introduction}\label{sec1}

Axial spondyloarthritis (axSpA) is a chronic, immune-mediated disease that predominantly affects the axial skeleton, particularly the sacroiliac joints (SIJ) and the spine. The disease spectrum encompasses both radiographic axSpA (also referred to as ankylosing spondylitis – AS), marked by visible structural damage to the SIJ on radiographs (normally fulfilling the radiographic criterion of the modified New York criteria for AS \cite{Linden1984Evaluation}, and non-radiographic axSpA, where such damage is absent \cite{Sieper2017Axial}. Detecting radiographic sacroiliitis still plays an essential role in diagnosing and classifying axSpA. According to the modified New York criteria, a sacroiliitis grade of $\geq$ 2 on both SIJs or a grade of 3-4 on one SIJ serves as the criterion for the presence of definite radiographic sacroiliitis \cite{Linden1984Evaluation}. While Magnetic Resonance Imaging (MRI) has proven instrumental in identifying active inflammatory and structural changes, particularly in early disease stages, conventional radiography retains certain advantages. It is more cost-effective, widely available, and more accessible in many regions of the world as compared to MRI. According to international recommendations focusing on imaging in SpA, radiography of sacroiliac joints is still recommended as the first imaging method if axSpA is suspected \cite{Mandl2015EULAR}. A critical limitation in detecting radiographic sacroiliitis, however, lies in the high inter- and intraobserver variability \cite{Yazici1987Observer}, with evidence indicating that specialized central readers outperform untrained counterparts \cite{vandenBerg2014}.
To address these issues, previous work has explored the utility of deep learning to detect radiographic sacroiliitis, as such a model could learn from the expert annotated training data and then be utilized to make consistent and accurate diagnoses even in medical centers without specialized human readers. The proposed convolutional neural network (CNN) achieved a performance that was comparable to human expert readers on standard pelvic radiographs \cite{Bressem2021Deep, Poddubnyy2021Detection}. Although axSpA primarily affects the SJI, a pelvic X-ray encompasses the entire pelvic area and surrounding soft tissue, introducing extraneous elements that reduce the signal-to-noise ratio of the image and potentially diminish the algorithm's effectiveness. However, extracting the SJI could increase the AI model's accuracy by eliminating irrelevant distractions and only focusing on the relevant anatomical areas. This concept of anatomical awareness is inspired by the process in which human readers assess radiographs: initially identifying the anatomical region of interest and subsequently assessing changes therein.
In this study, we aim to evaluate the efficacy of an anatomy-aware approach in the detection of sacroiliitis. Specifically, we compare the performance of deep learning models trained on cropped images, containing only the SIJ, to those trained on uncropped radiographs. Furthermore, we investigated whether deep learning models could detect changes in the image with a higher sensitivity than a human reader and thereby identify patients who are at a higher risk of progression to radiographic axSpA.

\section{Methods}\label{sec2}

\subsection{Cohort Description}
\begin{figure}[ht]
    \centering
    \includegraphics[width=\linewidth]{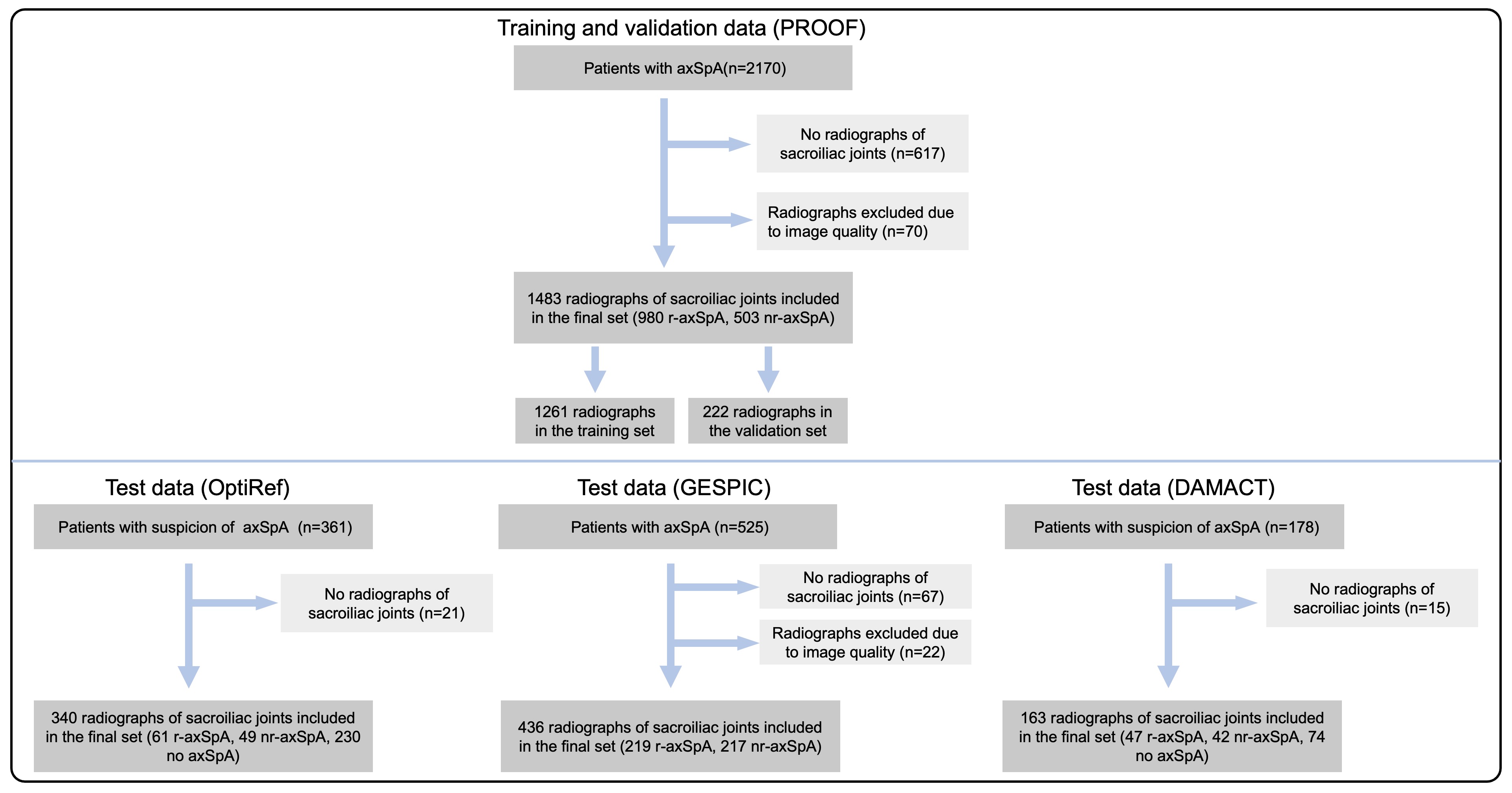}
    \caption{Flowchart for the selection of images from the PROOF dataset (training and validation) and GESPIC, DAMACT, and OptiRef datasets (testing). r-axSpA: radiographic axSpA, nr-axSpA: non-radiographic axSpA.}
    \label{fig:figure1}
\end{figure}

Imaging data from four different studies was used:
\begin{itemize}
    \item Patients With Axial Spondyloarthritis: Multicountry Registry of Clinical Characteristics (PROOF). PROOF is a continuing study carried out in clinical settings across 29 countries. It includes 2170 adult patients who have been diagnosed with axSpA (non-radiographic or radiographic) $\leq$ 1 year prior to study enrolment and fulfill the ASAS classification criteria for axSpA \cite{Poddubnyy2022Characteristics}. 1483 radiographs from PROOF (all of them diagnosed with axSpA, including radiographic and non-radiographic forms) were available as training data for our study. 
    \item German Spondyloarthritis Inception Cohort (GESPIC). GESPIC is a multicenter inception cohort study conducted in Germany and includes 525 patients with axSpA (non-radiographic or radiographic) \cite{Rudwaleit2009early}. In 436 patients (all of them diagnosed with axSpA, including radiographic and non-radiographic forms), radiographs of the sacroiliac joints were available for inference. 
    \item Identification of the Optimal Referral Strategy for Early Diagnosis of Axial Spondyloarthritis (OptiRef). OptiRef is a clinical study that includes a total of 361 patients who reported symptoms suggestive of axSpA \cite{Proft2020Comparison}. From this study, 340 images (110 diagnosed with axSpA, including radiographic and non-radiographic forms, and 230 without SpA) were available for inference.  
    \item Diagnostic Accuracy of MRI and CT in axSpA (DAMACT). DAMACT is a combined Dataset of the SacroIliac Magnetic Resonance Computed Tomograph (SIMACT) trial and the Virtual Non-Calcium / Susceptibility Weighted Imaging (VNCa/SWI) trial. The studies included a total of 178 patients with suspected axSpA \cite{Diekhoff2017Comparison, Deppe2021CT-like, Diekhoff2022Choose}. 163 radiographs (89 diagnosed with axSpA, including radiographic and non-radiographic forms, and 74 without SpA) of this dataset were available for inference.
\end{itemize}
Table \ref{tab:patient_characteristics} shows further information on the patients clinical characteristics and Figure \ref{fig:figure1} shows the patient flowchart.

\begin{table}[ht]
\centering
\caption{Patient Characteristics}
\label{tab:patient_characteristics}
\begin{tabular}{lcccc}
\hline
\addlinespace[0.5ex]

 & \textbf{PROOF} & \textbf{GESPIC} & \textbf{OPTIREF} & \textbf{DAMACT} \\ 
\textbf{Clinical Parameter} & N = 1,483 & N = 436 & N = 340 & N = 163 \\ 
\hline
\addlinespace[0.25ex]
\textbf{axSpA, n (\%)} & 1483 (100) & 436 (100) & 110 (32.4) & 89 (54.6) \\
\textbf{radiographic axSpA, n (\%)} & 980 (66.1) & 219 (50.2) & 61 (17.9) & 47 (28.8) \\
\textbf{non-radiographic axSpA, n (\%)} & 503 (33.9) & 217 (49.8) & 49 (14.4) & 42 (25.8) \\
\textbf{Male sex, n (\%)} & 941 (63.5) & 235 (53.9) & 167 (49.1) & 82 (50.3) \\
\textbf{Age, years, mean (S.D.)} & 34.6 (10.6) & 35.5 (10.2) & 38.1 (10.7) & 38.2 (10.6) \\
\textbf{Symptom duration, years, mean (S.D.)} & 4.7 (6.9) & 4.1 (2.7) & 8.4 (7.7) & - \\
\textbf{HLA-B27 positivity, n (\%)} & 789 (53.2) & 347 / 433 (80.1) & 153 / 334 (45.8) & 96/154 (62.3) \\
\textbf{Family history of SpA, n (\%)} & 278 (18.7) & 137 (31.4) & 50 / 302 (16.6) & 22/109 (20.2) \\
\textbf{Arthritis, n (\%)} & 475 (32.0) & 57 (13.1) & 65 (19.1) & 56 (34.4) \\
\textbf{Enthesitis, n (\%)} & 532 (35.9) & 91 (20.9) & 59 (17.4) & - \\
\textbf{Dactylitis, n (\%)} & 82 (5.5) & 9 (2.1) & 2 (0.6) & - \\
\textbf{Uveitis, n (\%)} & 145 (9.8) & 75 (17.2) & 31 (9.1) & 29/152 (19.1) \\
\textbf{Psoriasis, n (\%)} & 99 (6.7) & 42 (9.6) & 36 (10.6) & 22/152 (14.4) \\
\textbf{IBD, n (\%)} & 38 (2.6) & 12 (2.8) & 8 (2.4) & 10/137 (7.3) \\
\textbf{CRP, mg/L, mean (S.D.)} & 15.6 (23.1) & 11.3 (18.0) & 3.1 (5.9) & 5.4 (9.6) \\
\textbf{Elevated CRP, n (\%) }& 686 (46.3) & 188 / 423 (44.4) & 55 / 337 (16.3) & 20 (12.3) \\
\textbf{ASDAS-CRP, mean (S.D.)} & 2.9 (1.1) & 2.5 (1.0) & 2.4 (0.8) & - \\
\textbf{BASDAI, mean (S.D.)} & 4.5 (2.3) & 3.9 (2.1) & 4.4 (1.9) & 4.5 (1.8) \\
\textbf{BASFI, mean (S.D.)} & 3.3 (2.5) & 2.7 (2.3) & 2.6 (2.1) & - \\
\textbf{Treatment with NSAIDs, n (\%)} & 1,155 (77.9) & 295 (67.7) & - & - \\
\textbf{Treatment with systemic steroids, n (\%)} & 114 (7.7) & 37 (8.5) & - & - \\
\textbf{Treatment with csDMARDs, n (\%) }& 464 (31.3) & 96 (22.0) & - & - \\
\textbf{Treatment with a TNF inhibitor, n (\%) }& 198 (13.4) & 11 (2.5) & - & - \\
\hline
\end{tabular}
\smallskip

\small
ASDAS, Ankylosing Spondylitis Disease Activity Score; BASDAI, Bath Ankylosing Spondylitis Disease Activity Index; BASFI, Bath Ankylosing Spondylitis Functional Index; CRP, C reactive protein; DMARDs, disease-modifying antirheumatic drugs; HLA-B27, human leukocyte antigen B27; IBD, inflammatory bowel disease; NSAID, non-steroidal anti-inflammatory drugs; S.D., standard deviation; SpA, spondyloarthritis; TNFi, tumor necrosis factor alpha
\centering
\end{table}

\subsection{Label generation}
Radiographs of the sacroiliac joints were collected, anonymized, and subsequently graded by trained and calibrated readers using the modified New York criteria. For each image, grading was performed by multiple expert readers. Please refer to the Appendix for a detailed description of the grading process.

\subsection{Datasets and pre-processing}
The images used in the training and testing phases are sourced from different datasets, as described above. The Pelvis and Sacrum were manually segmented on images in the training dataset using 3D Slicer and MONAI Label \cite{Fedorov20123D, diaz-pinto_monai_2023}. These were used to train the Segmentation Model. In the segmentations generated by the model the sacrum labels were dilated and the resulting intersection with the pelvis was used to demark the SIJ, around which bounding boxes were defined. The boxes served as the basis for creating cropped images that show only the SIJ and the area in between. The same preprocessing was applied to the full radiographs and the cropped images to ensure consistency in the data that fed into the different models. 

The PROOF Dataset was used to train the model and was therefore randomly split into a training dataset (1261 images, 85\%) and a validation dataset (222 images, 15\%). The GESPIC, OptiRef, and DAMACT datasets were each used as independent test datasets.

\begin{figure}[ht]
    \centering
    \includegraphics[width=\linewidth]{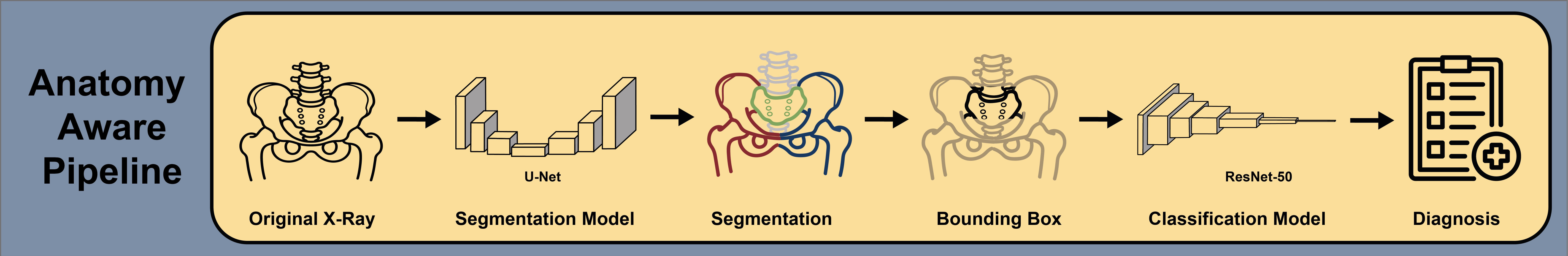}
    \caption{Illustration of the anatomy-aware processing pipeline}
    \label{fig:figure2}
\end{figure}

\subsection{Model training}
For Segmentation, we used a CNN based on the U-Net architecture \cite{ronneberger_u-net_2015}. For preprocessing, all images were resized to 512x512 pixels and their intensity was normalized. The weighted sum of the Dice loss and Cross Entropy Loss was used as the loss function. The training was performed using the AdamW optimizer algorithm and the one-cycle learning rate policy as described by Smith et al. \cite{loshchilov_decoupled_2019, LeslieSuper-convergence}. The training data was augmented using random transformations, such as cropping, rotation up to 10°, shearing, intensity and contrast shifting, and Gaussian noise. 

For Classification, we utilized a CNN based on the ResNet-50 architecture, pre-trained on the ImageNet-1k dataset \cite{He2016Deep}. For preprocessing, all images underwent resizing and cropping to uniform dimensions. A Contrast Limited Adaptive Histogram Equalization (CLAHE) filter was further applied to normalize intensity and improve contrast \cite{Qiu2017Automatic}.  To augment the training data, we introduced a multitude of random transformations - Comprising random flipping, random rotating up to 10°, random zooming by $\pm$ 10\%, and random shearing. Cross Entropy loss with label smoothing was used as the loss function. Again, the AdamW optimizer algorithm and the one-cycle learning rate policy were used. Additionally, we implemented discriminative learning rates, in which the learning rate increased for deeper layers. Mixup was further used to reduce overfitting \cite{zhang_mixup_2018, szegedy_rethinking_2016}. A progressive resizing approach was implemented, training first on images with sizes of 106 x 158 pixels and then increasing the resolution through 208 x 314, 312 x 472 up to 416 x 628 pixels after every 25, 25, 30, and 40 epochs of training respectively. Both full radiographs and SIJ-cropped images were employed to train two separate but fully comparable models. These models' performance was then examined across different testing datasets. A comprehensive detailing of the training regimens is available in the appendix. To validate the model's focus on clinically relevant features, Gradient-weighted Class Activation Mapping (Grad-CAM) was employed \cite{Selvaraju2020Grad-CAM:}. The training was performed on a dedicated Ubuntu workstation with two Nvidia GeForce RTX 2080 Ti graphics cards and on a GPU Server of the Scientific Computing Cluster of the Charité Universitätsmedizin Berlin using two Nvidia A100 graphics cards. The training was performed using PyTorch (Version 1.13.0) with PyTorch Lightning (Version 1.8.6) and MONAI (Version 1.1.0) \cite{Falcon2019PyTorch, cardoso_monai_2022}. All code and the anatomy aware model are available as fully open source at: https://github.com/FJDorfner/Anatomy-Aware-Classification-axSpA.

\subsection{Follow-up analysis}
For the GESPIC dataset, follow-up data was available for a total of 311 patients at different time points. The most data points were available for the 2-year follow-up, which includes 251 patients. Out of these, 135 were initially classified as not having radiographic sacroiliitis. Odds ratios were calculated to compare the proportion of patients with a confident false positive prediction at the baseline timepoint to the total patients without radiographic axSpA at baseline for both models. Predictions were considered confident for probabilities $>$ 0.7. 

\subsection{Statistical analysis}
Metrics for training and validation/testing were computed using TorchMetrics \cite{Detlefsen2022TorchMetrics}. The Python packages NumPy (Version 1.22.2), pandas (Version 1.4.4), scikit-learn (Version 0.24.2), matplotlib (Version 3.7.1), and seaborn (Version 0.12.2) were used for data analysis and visualization \cite{Harris2020Array, Pedregosa2011Scikit-learn:, mckinney_data_2010, Hunter2023Matplotlib:, Waskom2021seaborn:}. The optimal cut-off value - the value above which model probabilities are considered to be positive - was determined on the validation dataset by repeatedly calculating the accuracy for different cut-offs. The confidence of predictions was visualized. Receiver Operating Characteristic Curves (ROC) were plotted, and the Area Under the Receiver Operating Characteristic Curve (AUC) was calculated. Confusion matrices were generated to further illustrate the model's effectiveness. Given data imbalances across datasets, balanced accuracy, sensitivity, and specificity were calculated. 95\% confidence intervals were estimated using bootstrapping with 1000 repetitions. DeLong’s algorithm was used to compare the AUC between the two models for each dataset \cite{Sun2014Fast}. McNemar’s test was used to compare model performance on the follow-up data. A p-value $<$0.05 was considered statistically significant.

\subsection{Ethics approval}
PROOF, GESPIC, OptiRef, SIMACT, and VNCa/SWI studies were approved by the local ethics committees of each study center in accordance with the local laws and regulations and were conducted in accordance with the Declaration of Helsinki and Good Clinical Practice. Written informed consent was obtained from all patients.

\section{Results}\label{sec3}
Expert readers identified radiographic sacroiliitis in 835 (66\%) images from the PROOF training dataset (n=1261) and in 145 (65\%) from the PROOF validation dataset (n=222). Regarding the independent test datasets, radiographic sacroiliitis was present in 219 (50\%) of 436 GESPIC images, 47 (29\%) of 163 DAMACT images, and 61 (18\%) of 340 OptiRef images. Classification performance of the anatomy-aware and standard model was compared on the validation and the test datasets. 

\subsection{Validation Dataset Model Comparison}
Both the standard model and the anatomy-aware models exhibited strong performance on the PROOF validation dataset. The AUC was 0.888 for the standard model and 0.9 for the anatomy-aware model. 
For both models, the cut-off value that results in the highest validation accuracy was calculated. This cut-off value was 0.69 for the standard model and 0.59 for the anatomy-aware model. These cut-offs were then used for all further predictions with the respective model on the test datasets. For the standard model, this resulted in an accuracy of 0.826, specificity of 0.844, and sensitivity of 0.807. For the anatomy-aware model, the cut-off value resulted in an accuracy of 0.836, specificity of 0.844, and sensitivity of 0.828. ROC curves and confusion matrices for the validation dataset are provided in the supplementary materials. 

\subsection{Test Dataset Model Comparison}
Performance disparities between the models were dataset-dependent. Table \ref{tab:model_performance} summarizes the classification metrics for both models and each of the independent test datasets. In GESPIC, the standard model attained an AUC of 0.853, whereas the anatomy-aware model reached 0.899. For DAMACT, AUC scores were 0.817 for the standard model and 0.846 for the anatomy-aware model. For OptiRef, the AUC scores were 0.947 for the standard model and 0.957 for the anatomy-aware model. The difference between the two models for GESPIC was statistically significant (p=0.0047). For the DAMACT and OptiRef data, the difference between the two models was not statistically significant (p = 0.18 and 0.22, respectively).
Confusion matrices for each dataset and model can be found in the Supplements.

\begin{table}[H]
\centering
\caption{Classification performance for the standard and the anatomy-aware model on three different independent test datasets}
\label{tab:model_performance}
\begin{tabular}{lcc}
\hline
\addlinespace[0.5ex]
 & \textbf{Standard Model} & \textbf{Anatomy-Aware Model} \\ 
\addlinespace[0.5ex]
\hline
\addlinespace[0.5ex]
\textbf{GESPIC Test Dataset} \\[0.2ex]
AUC & 0.853 (95\% CI: 0.817, 0.887) & 0.899 (95\% CI: 0.866, 0.928) \\[0.5ex]
Balanced Accuracy & 0.770 (95\% CI: 0.731, 0.808) & 0.821 (95\% CI: 0.786, 0.853) \\[0.5ex]
Sensitivity & 0.886 (95\% CI: 0.847, 0.927) & 0.941 (95\% CI: 0.908, 0.969) \\[0.5ex]
Specificity & 0.654 (95\% CI: 0.587, 0.716) & 0.700 (95\% CI: 0.642, 0.760) \\[0.5ex]
\hline
\addlinespace[0.5ex]
\textbf{DAMACT Test Dataset} \\[0.5ex]
AUC & 0.817 (95\% CI: 0.740, 0.884) & 0.846 (95\% CI: 0.779, 0.997) \\[0.5ex]
Balanced Accuracy & 0.724 (95\% CI: 0.647, 0.801) & 0.744 (95\% CI: 0.670, 0.805) \\[0.5ex]
Sensitivity & 0.681 (95\% CI: 0.550, 0.818) & 0.851 (95\% CI: 0.739, 0.945) \\[0.5ex]
Specificity & 0.767 (95\% CI: 0.687, 0.841) & 0.638 (95\% CI: 0.548, 0.721) \\[0.5ex]
\hline
\addlinespace[0.5ex]
\textbf{OptiRef Test Dataset} \\[0.5ex]
AUC & 0.947 (95\% CI: 0.921, 0.968) & 0.957 (95\% CI: 0.926, 0.979) \\[0.5ex]
Balanced Accuracy & 0.850 (95\% CI: 0.793, 0.903) & 0.906 (95\% CI: 0.868, 0.942) \\[0.5ex]
Sensitivity & 0.754 (95\% CI: 0.638, 0.860) & 0.934 (95\% CI: 0.870, 0.986) \\[0.5ex]
Specificity & 0.946 (95\% CI: 0.918, 0.971) & 0.878 (95\% CI: 0.839, 0.910) \\[0.5ex]
\hline
\end{tabular}
\end{table}

\begin{figure}[ht]
    \centering
    \includegraphics[width=\linewidth]{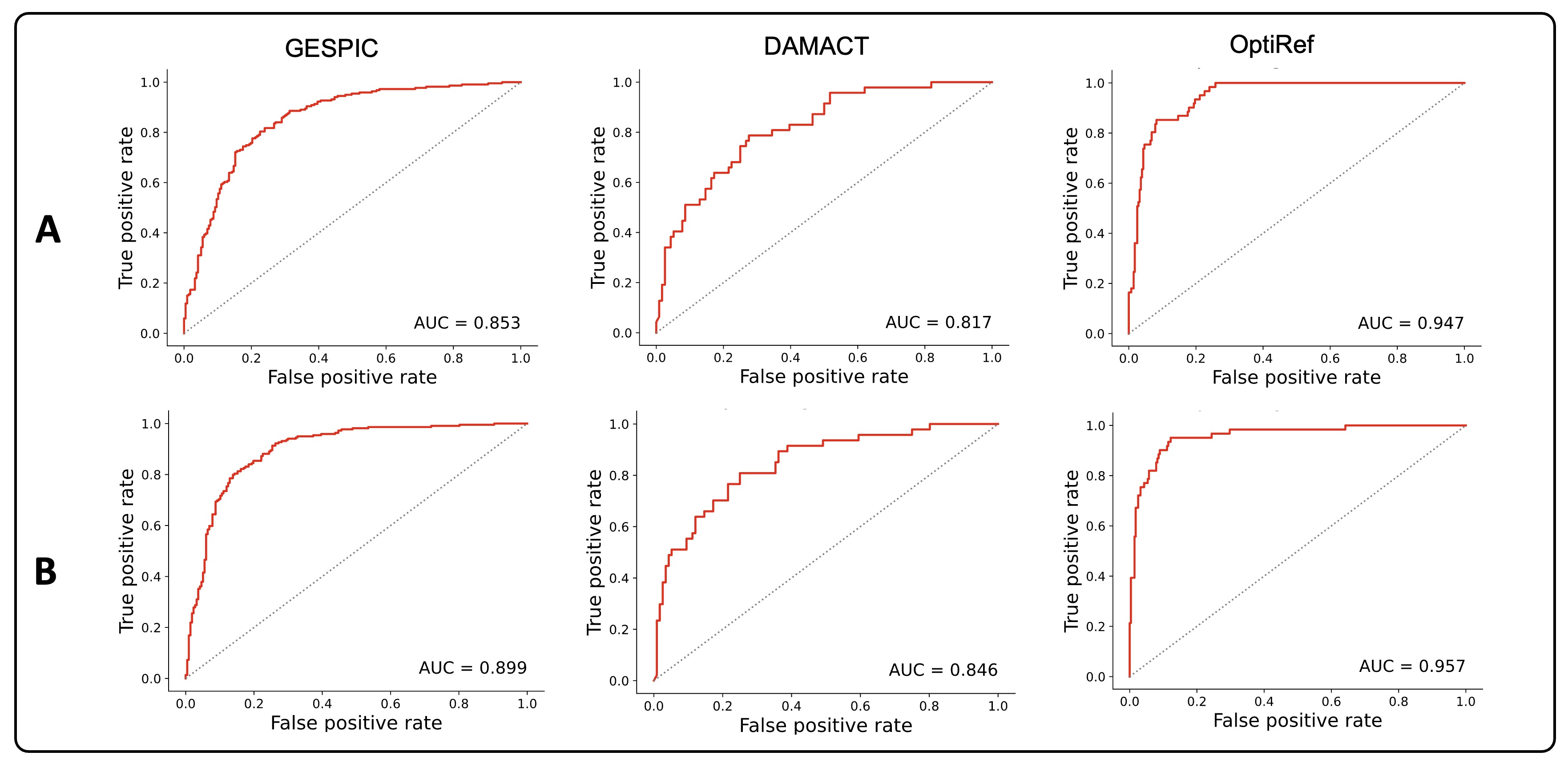}
    \caption{Receiver operating characteristics curves and area under the curves for the standard model (A) and the anatomy-aware model (B).}
    \label{fig:figure3}
\end{figure}

\subsection{Investigation of Model Activations }
A comparison of the Grad-CAMs for both models revealed that the anatomy-aware model exhibited more targeted activations, specifically focusing on the SIJ. This attention to diagnostically relevant regions aligns with the New York classification criteria, which categorize patients as positive for radiographic sacroiliitis if they exhibit a score greater than or equal to grade II bilaterally or greater than or equal to grade III unilaterally. 

\begin{figure}[H]
    \centering
    \includegraphics[width=\linewidth]{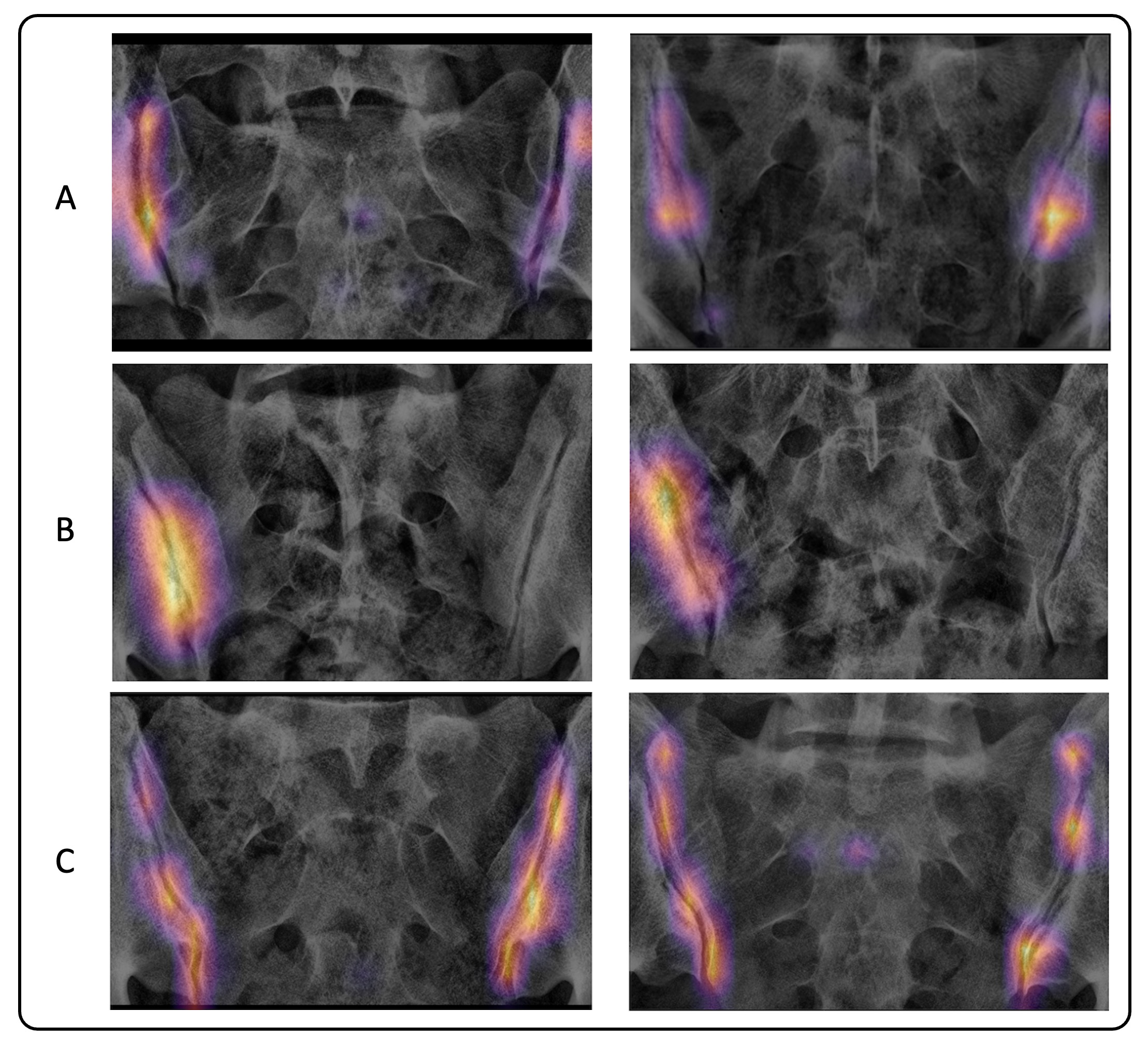}
    \caption{Grad-CAMs of the anatomy-aware model for predictions from the test dataset. A:  mild sacroiliitis is present on both sides, and the model's activations are equally focused on both sides; B: sacroiliitis of grade $\geq$3 present, and the model has based its decision for radiographic axSpA on only one joint and especially on the lesion; C: images without radiographic axSpA, in those the model is focused equally on both sides and the full length of the SIJ.}
    \label{fig:figure4}
\end{figure}

\subsection{Evaluating Model Confidence Across Datasets}
We performed an analysis of the models' output probabilities across various datasets to evaluate their confidence in predictions. The output probabilities - values before applying the cut-off for binary classification - are closer to ground truth (either 0 or 1) when the model is more confident. An ideal model would provide probabilities that align closely with the ground truth.

In the GESPIC dataset, both models were more confident in predicting positive cases for radiographic axSpA than in negative cases. For the DAMACT dataset, both models show similar confidence values for the positive and negative class. In the OptiRef dataset, both models showed a higher degree of confidence predicting negative cases of non-radiographic axSpA.  

\begin{figure}[H]
    \centering
    \includegraphics[width=\linewidth]{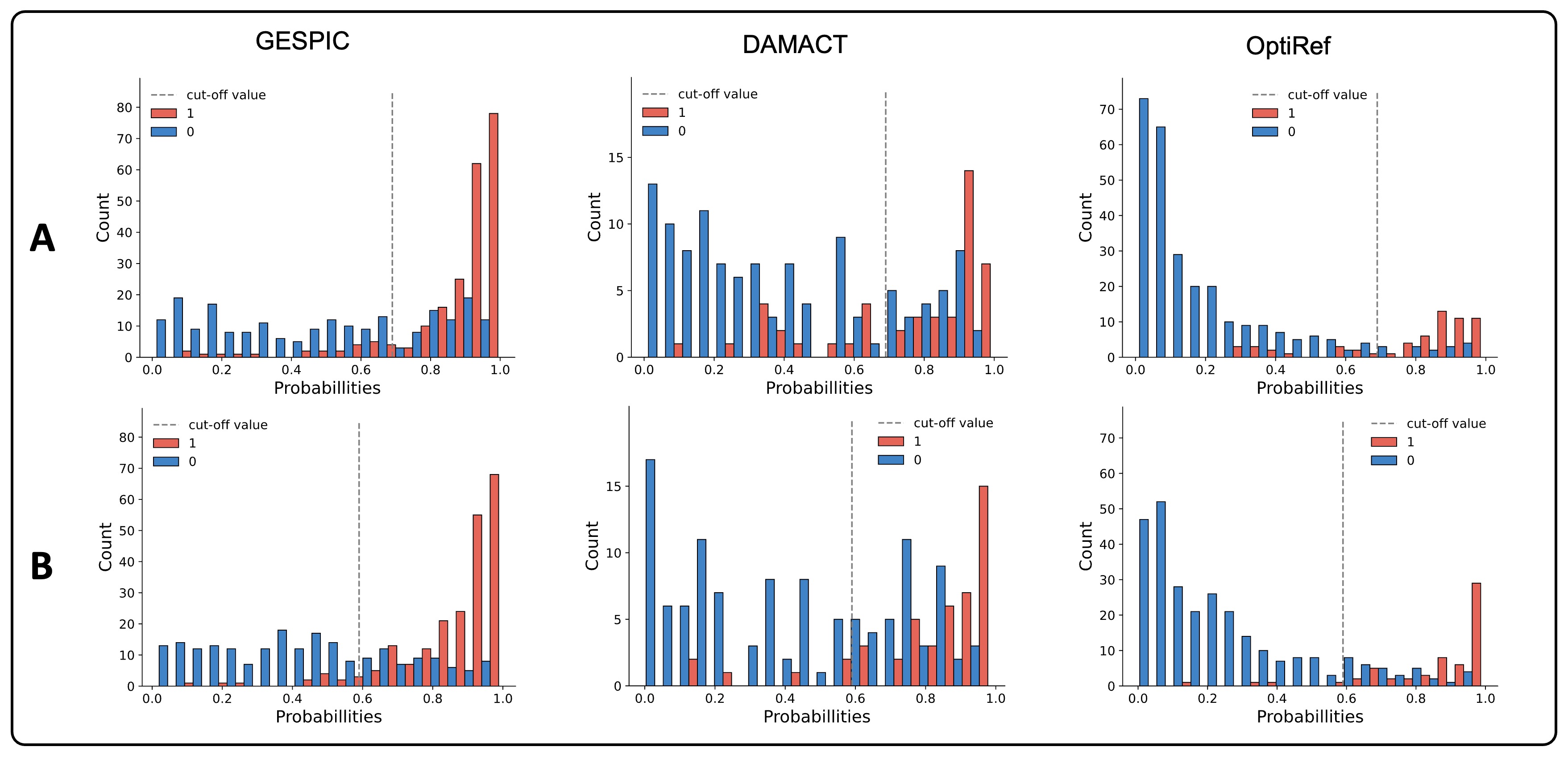}
    \caption{Probabilities for the standard model (A) and anatomy-aware model (B). 1 refers to r-axSpA and 0 to nr-axSpA. The bars are colored according to the respective ground truth.}
    \label{fig:figure5}
\end{figure}

\subsection{Identifying High-Risk Patients}
The ability of deep learning models to detect subtle features in medical images can be potentially leveraged for identifying patients at a higher risk of disease progression (in our case - of development of definite radiographic sacroiliitis). To test the hypothesis that the model is able to detect non-radiographic patients with a high probability of radiographic sacroiliitis development at baseline later on, we utilized the data of the 2-year follow-up in the GESPIC dataset. This included a total of 251 patients, of which 135 were initially classified as non-radiographic axSpA. 25 of these patients (18.5\%) progressed to radiographic axSpA after two years.
We analyzed the predictions of both models for those patients at baseline. The anatomy-aware model had 30 confident false positive predictions, of which 12 (40\%) had a progression within 2 years. This equals an odds ratio of 2.16 (95\% CI: 1.19, 3.86) for progression after two years. The standard model had 48 confident false positive predictions, of which 11 (22.9\%) had a progression within two years, which resulted in an odds ratio of 1.24 (95\% CI: 0.60, 2.34). The difference in model performance when considering the labels at the 2-year follow-up was statistically significant (p=0.0002).

\section{Discussion}\label{sec4}
This study shows incorporating anatomical awareness in artificial intelligence algorithms augments their capacity to generalize, thereby enhancing their utility and reliability in predicting definite radiographic sacroiliitis on unseen data. Both the standard and anatomy-aware models exhibited comparable learning abilities from the training data but substantial performance differences on the different external test datasets used. Particularly, the anatomy-aware model showed a higher consistency across datasets, with minimal fluctuations in performance metrics, despite changes in disease prevalence and severity. This improved performance also manifested in more targeted model activations, as demonstrated by the Grad-CAMs. Here, the anatomy-aware model displayed dual advantages. Firstly, it avoided distractions from other anatomical structures, such as the pubic symphysis or hip joint, that often sidetrack the standard model. Secondly, even when the sacrum appears in the frame, the anatomy-aware model remains committed to focusing on the SIJ. This suggests that the advantage of the anatomy-aware model lies not only in eliminating distracting joints in preprocessing, but that the anatomy-aware model has also learned to better locate and assess the SIJ.

\begin{figure}[H]
    \centering
    \includegraphics[width=\linewidth]{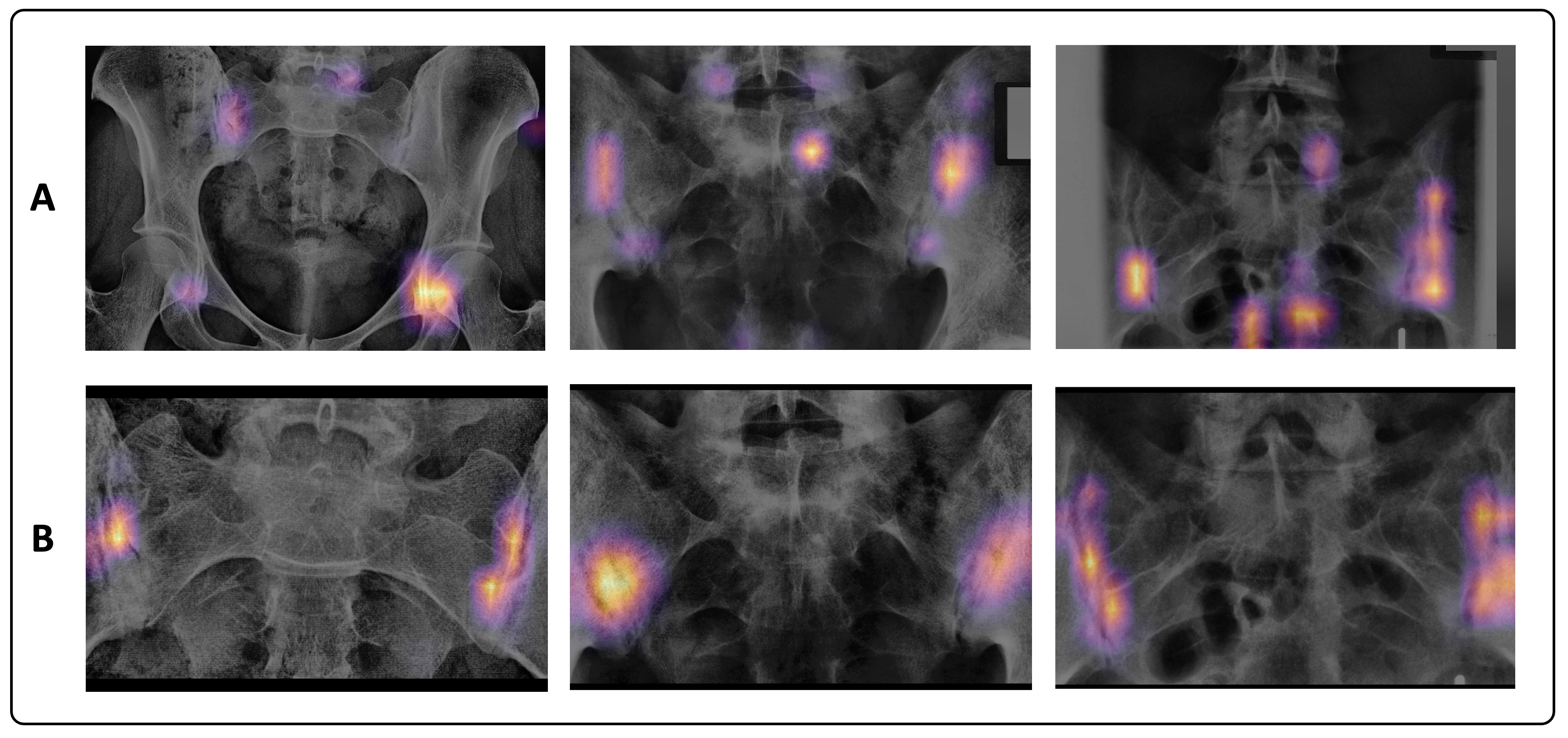}
    \caption{Grad-CAMs for example images from GESPIC as output by the standard model (A) and the anatomy-aware model (B).}
    \label{fig:figure6}
\end{figure}

Moreover, the anatomy-aware model shows a nuanced capability to identify patients who may be on a trajectory toward disease progression in terms of definite radiographic sacroiliitis development. Drawing from the GESPIC dataset, the model expressed high confidence in diagnosing radiographic axSpA in a specific subset of patients who were initially classified as not having definite radiographic sacroiliitis by human experts. Specifically, these patients were twice as likely to progress to radiographic axSpA within two years. However, it is not clear if the model presciently forecasted the development of radiographic sacroiliitis or if it rather more accurately discerned early-stage markers that human readers might deem inconclusive for a formal diagnosis. Independent of this, the model's confident 'false positives' could serve as an actionable flag for clinicians, guiding them to prioritize these patients for more intensive surveillance. This could make the model a valuable aid in identifying patients that could most benefit from a proactive and focused clinical follow-up, thereby enhancing patient management strategies in a nuanced manner. 
Compared to previous studies \cite{Bressem2021Deep, Poddubnyy2021Detection}, our work has two key advantages: Firstly, this study utilizes automatic pre-processing, with uniform settings for all images, instead of manually adjusting grayscale levels for each image as done in \cite{Bressem2021Deep}, which is more representative of a real-world clinical application. Secondly, while Üreten et al. also cropped the pelvic radiographs to include only the SIJ before training a CNN \cite{Ureten2023Deep}. They evaluated the model performance on a small test dataset obtained from one site. In comparison, this study used multiple independent test datasets, providing a better picture of the model's ability to generalize to images taken from patients with different clinical characteristics. Furthermore, the direct comparison between the standard and the anatomy-aware model allowed the assessment of the impact that anatomical awareness has on the generalizability of the model. To the best of our knowledge, there were no other studies investigating the concept of anatomical awareness in pelvic radiographs at the time of writing.
There are also some limitations to this study. The first Follow-Up examination that was available for this study was conducted after two years. It is therefore unknown when exactly during this two-year period the progression could have been diagnosed or if additional patients would progress after the two years. It seems probable that the progression becomes harder to predict the further it lies in the future. More data points with smaller intervals between examinations could provide a clearer picture of the model's ability to identify high-risk patients. 

\subsection*{Conclusion}
The incorporation of anatomical awareness can improve the diagnostic performance and generalizability of deep neural networks in detecting radiographic sacroiliitis. Additionally, neural networks may help identify patients who are likely to progress to radiographic axSpA. 

\FloatBarrier

\section*{Acknowledgements}
The authors acknowledge the Scientific Computing of the IT Division at the Charité - Universitätsmedizin Berlin for providing computational resources that have contributed to the research results reported in this paper. URL: 
https://www.charite.de/en/research/research\_support\_services/research\_infrastructure/science\_it/\#c30646061
KKB is grateful for his participation in the BIH Charité Digital Clinician Scientist Program funded by the Charité–Universitätsmedizin Berlin and the Berlin Institute of Health.

\newpage
\bibliographystyle{unsrtnat}
\bibliography{references.bib}  

\newpage
\section*{Supplemental Material}\label{sec6}

\subsection*{Label generation for the individual datasets}
Label generation for the individual datasets
For the PROOF dataset, images were first assessed by the local readers, then by a central reader, who was blinded to the results of the previous assessment. In the case of disagreement between the local and central reader, the radiograph was further evaluated by a second central reader. The final decision on the presence of definite radiographic sacroiliitis was made based on the decision of two out of the three readers. 
For the GESPIC dataset, all radiographs were scored independently by two trained and calibrated central readers.
For the OptiRef dataset, the radiographs were scored as the consensus of a rheumatologist and radiologist on the presence of definite radiographic sacroiliitis.
In SIMACT and VnCa, all radiographs were scored independently by three readers, who were blinded to other results. The agreement of two out of the three readers was used as the final score. 

\subsection*{Model Training Procedure}
The segmentation model was trained without pre-trained weights, for 500 epochs, using a learning rate of $1\times10^{-3}$  and a batch size of 8. During training, the mean Dice Coefficient on the validation dataset was monitored, and the model was saved on every improvement in the metric.  
For the classification model, first, only the classification head of the model was trained for 15 epochs on the smallest image size while the remaining layers were frozen. During training, the Matthews Correlation Coefficient on the validation dataset was monitored, and the model saved on every improvement in the metric. The best-performing model was then loaded and trained on all layers for another 15 epochs. This process was repeated for all image sizes with varying hyperparameters. The batch size was 32 for the image size 416x628 and 64 for all other image sizes. The total training time was two hours. The optimal values for the initial learning rate, the weight decay, the level of mixup, the coefficient discriminative learning rate, the number of training epochs, the batch size, and augmentation transforms were determined by testing a wide range of possible values and selecting the hyperparameters that produce the best classification accuracy. PyTorch Lightnings integration of mixed precision training was used to speed up the training process and reduce memory requirements.

\begin{figure}[H]
    \centering
    \includegraphics[width=\linewidth]{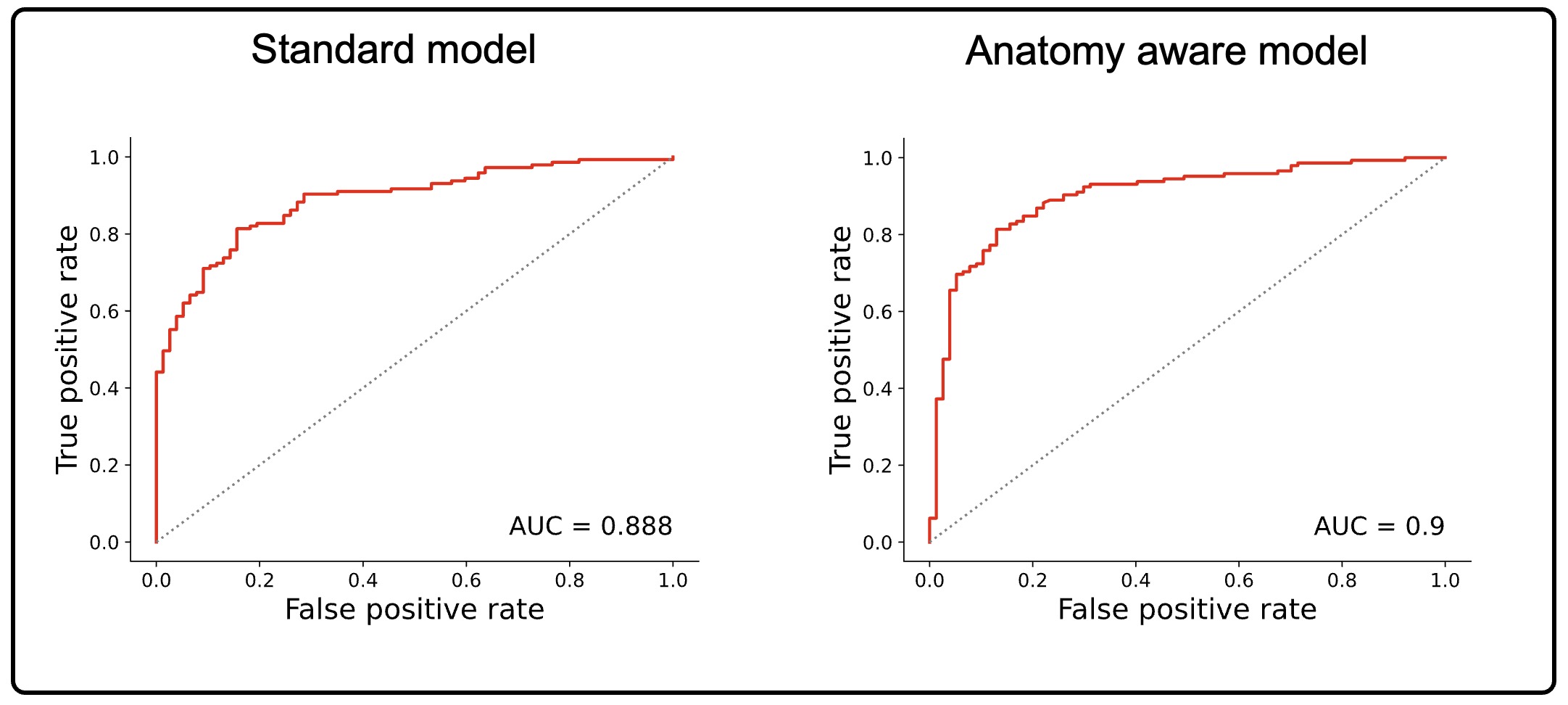}
    \caption{Receiver operating characteristics curves and area under the curves for the standard model and the anatomy-aware model, both on the validation dataset.}
    \label{fig:figure7}
\end{figure}

\begin{figure}[H]
    \centering
    \includegraphics[width=\linewidth]{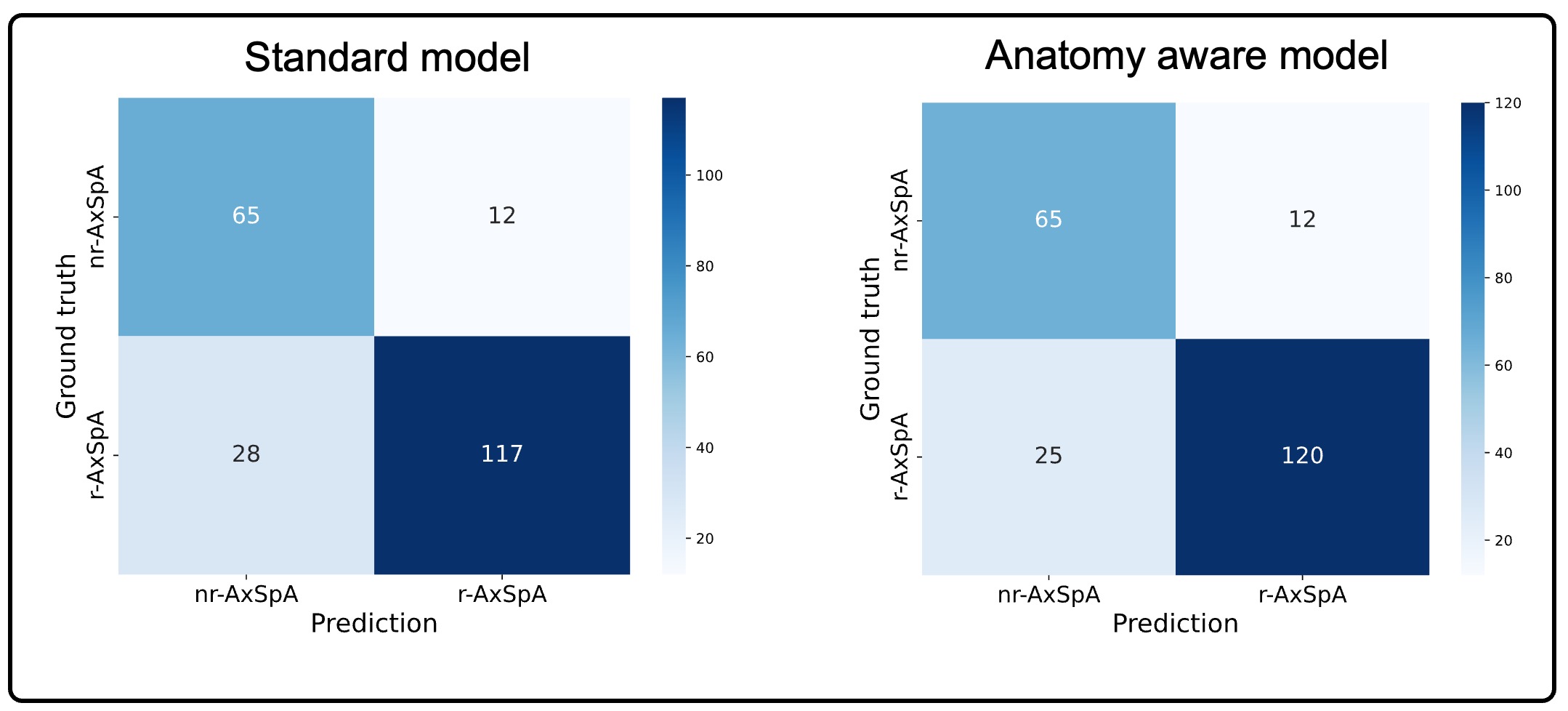}
    \caption{Confusion matrices for the standard model and the anatomy-aware model, both on the validation dataset.}
    \label{fig:figure8}
\end{figure}

\begin{figure}[H]
    \centering
    \includegraphics[width=\linewidth]{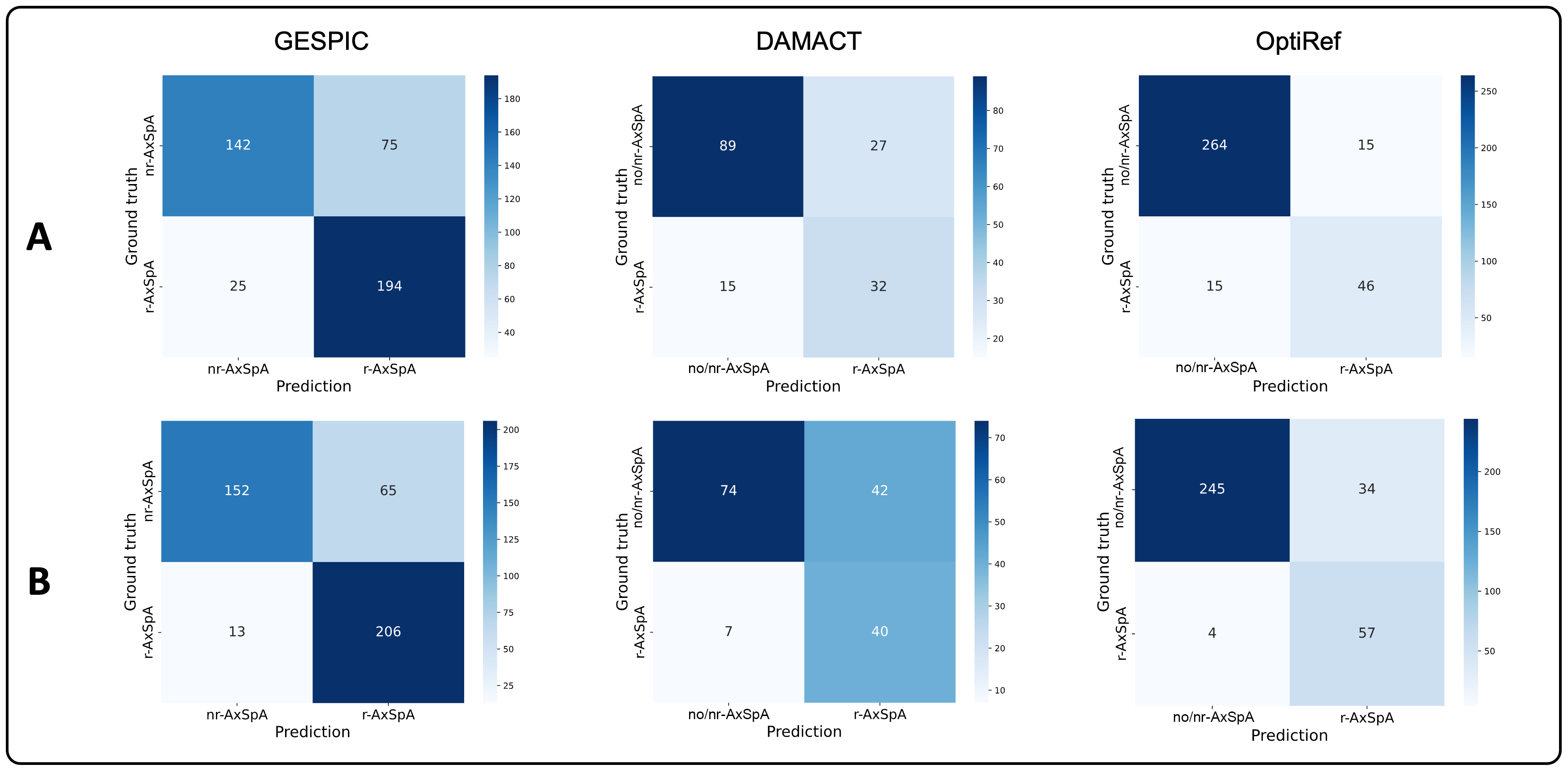}
    \caption{Confusion matrices for the standard model (A) and the anatomy-aware model (B).}
    \label{fig:figure9}
    
\end{figure}
\begin{figure}[H]
    \centering
    \includegraphics[width=\linewidth]{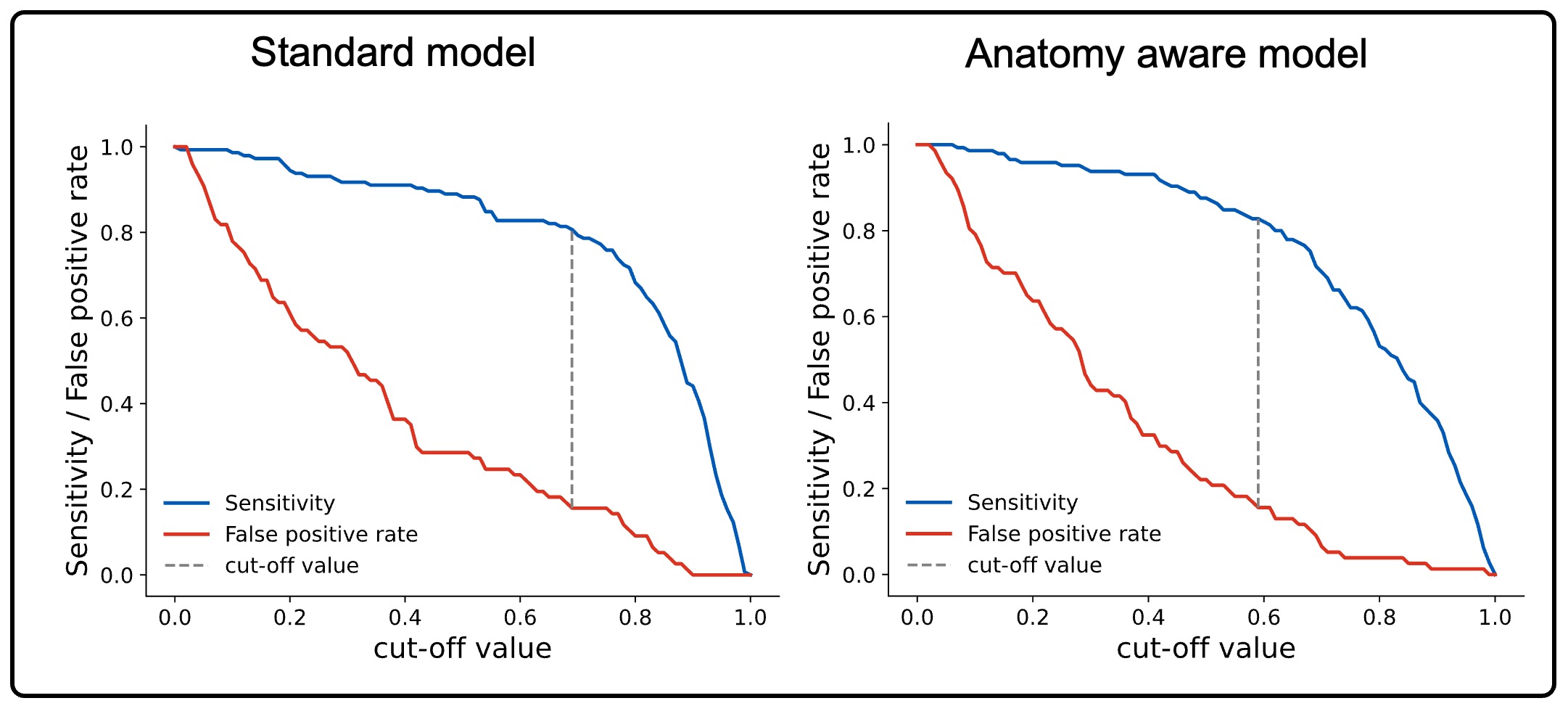}
    \caption{Sensitivity and 1- Specificity (False positive rate) for the standard model and the anatomy-aware model, both on the validation dataset. The Cut-off is 0.69 for the standard model and 0.59 for the anatomy-aware model.}
    \label{fig:figure10}
\end{figure}

\begin{figure}[H]
    \centering
    \includegraphics[width=\linewidth]{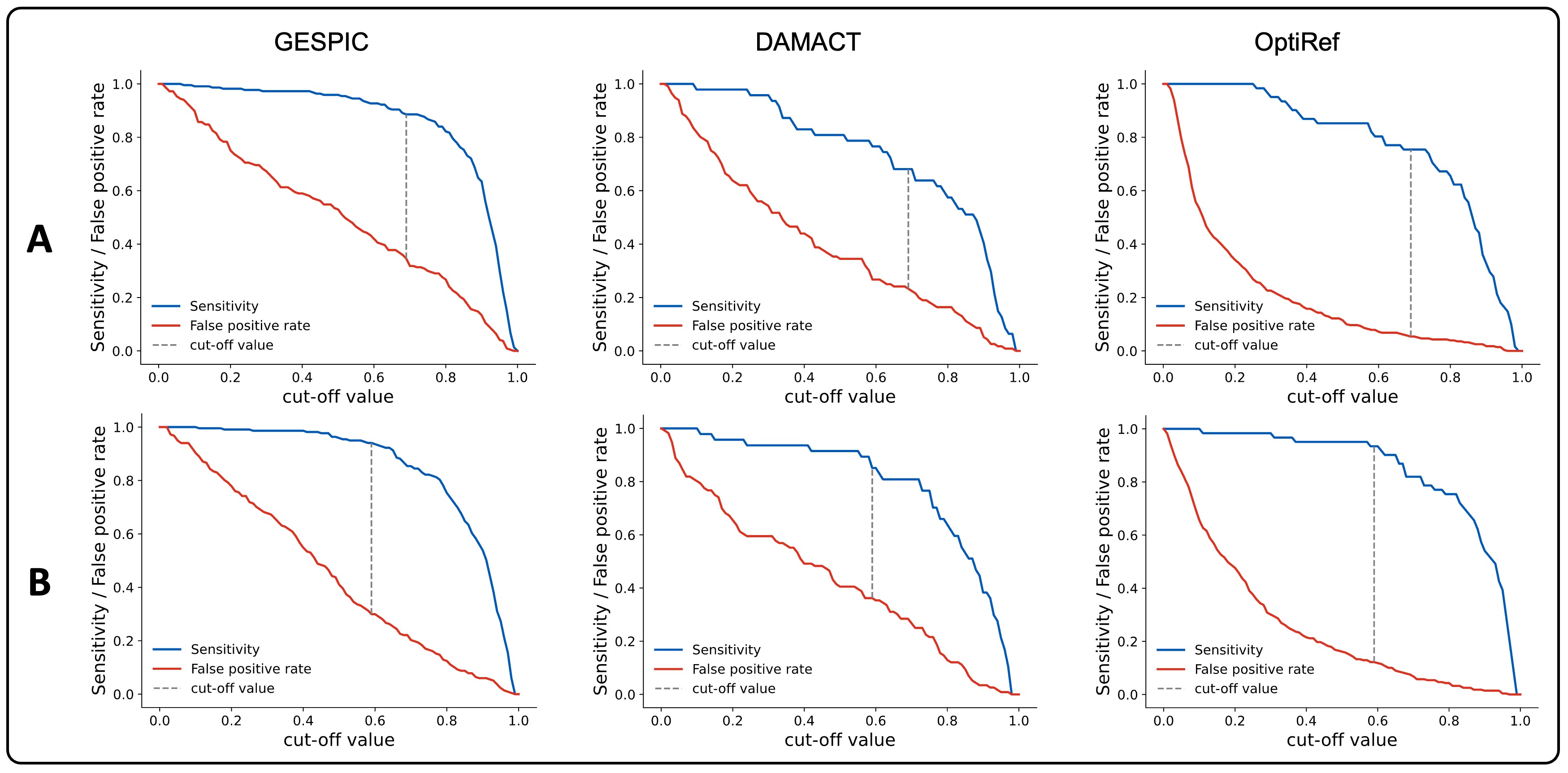}
    \caption{Sensitivity and 1- Specificity (False positive rate) for the standard model (A) and the anatomy-aware model (B). The Cut-off is 0.69 for the standard model and 0.59 for the anatomy-aware model.}
    \label{fig:figure11}
\end{figure}
\end{document}